\documentclass[10pt, a4paper]{article}
\usepackage{lrec2022} 
\usepackage{multibib}
\newcites{languageresource}{Language Resources}
\usepackage{graphicx}
\usepackage{tabularx}
\usepackage{soul}
\usepackage{titlesec}
\titleformat{\section}{\normalfont\large\bfseries\center}{\thesection.}{1em}{}
\titleformat{\subsection}{\normalfont\SmallTitleFont\bfseries\raggedright}{\thesubsection.}{1em}{}
\titleformat{\subsubsection}{\normalfont\normalsize\bfseries\raggedright}{\thesubsubsection.}{1em}{}
\renewcommand\thesection{\arabic{section}}
\renewcommand\thesubsection{\thesection.\arabic{subsection}}
\renewcommand\thesubsubsection{\thesubsection.\arabic{subsubsection}}

\usepackage{epstopdf}
\usepackage[utf8]{inputenc}

\usepackage{hyperref}
\usepackage{xstring}

\usepackage{color}

\usepackage{microtype}
\usepackage{amsmath,amssymb,amsfonts}
\usepackage{algorithmic}
\usepackage{textcomp}
\usepackage{xcolor}
\usepackage{cleveref}
\usepackage{fixltx2e}
 \usepackage{multirow}
  \usepackage{makecell}
\usepackage[absolute]{textpos}

\title{Towards Evaluation of Cross-document Coreference Resolution Models Using Datasets with Diverse Annotation Schemes}

\name{Anastasia Zhukova$^1$, Felix Hamborg$^2$, Bela Gipp$^{1,3}$} 

\address{$^1$ University of Wuppertal \\
         Gaußstraße 20, 42119 Wuppertal, Germany \\
         \{last\}@uni-wuppertal.de\\
         \\
         $^2$ University of Konstanz\\
         Universit\"atsstraße 10, 78464 Konstanz, Germany\\
         felix.hamborg@uni-konstanz.de\\
         \\
         $^3$ University of G\"ottingen\\
         Wilhelmsplatz 1, 37073 G\"ottingen, Germany\\
         gipp@uni-goettingen.de\\
         }

\abstract{
Established cross-document coreference resolution (CDCR) datasets contain event-centric coreference chains of events and entities with identity relations. These datasets establish strict definitions of the coreference relations across related tests but typically ignore anaphora with more vague context-dependent loose coreference relations. In this paper, we qualitatively and quantitatively compare the annotation schemes of ECB+, a CDCR dataset with identity coreference relations, and NewsWCL50, a CDCR dataset with a mix of loose context-dependent and strict coreference relations. We propose a phrasing diversity metric (PD) that encounters for the diversity of full phrases unlike the previously proposed metrics and allows to evaluate lexical diversity of the CDCR datasets in a higher precision. The analysis shows that coreference chains of NewsWCL50 are more lexically diverse than those of ECB+ but annotating of NewsWCL50 leads to the lower inter-coder reliability. We discuss the different tasks that both CDCR datasets create for the CDCR models, i.e., lexical disambiguation and lexical diversity. Finally, to ensure generalizability of the CDCR models, we propose a direction for CDCR evaluation that combines CDCR datasets with multiple annotation schemes that focus of various properties of the coreference chains. 
 \\ \newline 
 }

\begin{document}
\begin{textblock}{10}(2,1)
\noindent\small BibTeX: \url{https://aclanthology.org/2022.lrec-1.522/}
\end{textblock}

\maketitleabstract

\section{Introduction}
Cross-document coreference resolution (CDCR) is a set of techniques that aims to resolve mentions of events and entities across a set of related documents. CDCR is employed as an essential analysis component in a broad spectrum of use cases, e.g., to identify potential targets in sentiment analysis or as a part of discourse interpretation. 

Although CDCR research has gained attention, the annotation schemes and their corresponding datasets have only infrequently explored the mix of identity and lose coreference relations and lexical diversity of the annotated chains of mentions. For example, resolution of identity relations, i.e., coreference resolution, and resolution of more lose relations, i.e., bridging, are typically split into two separate tasks \cite{kobayashi-ng-2020-bridging}. Resolution and evaluation of entity and event mentions of the two types of relations remain a research gap in CDCR research. 

This paper explores the qualitative and quantitative characteristics of two CDCR annotation schemes that annotate both events and entities. (1) ECB+, a state-of-the-art event-centric CDCR dataset \cite{cybulska-vossen-2014-using} that annotates mentions with identity coreference relations. (2) NewsWCL50, an experimental concept-centric dataset for identification of semantic concepts, each containing mentions with a mix of identity, loose coreference, and bridging relations, e.g., context-dependent anaphora, synonyms, metonyms, and meronyms \cite{Hamborg2019a}. We propose a phrasing diversity metric (PD) that describes the variation of wording in annotated coreference chains and measures the lexical complexity of CDCR datasets. Unlike the number of unique lemmas, i.e., a previously used metric for lexical diversity \cite{Eirew2021}, the proposed PD allows capturing higher lexical variation in coreference chains. We discuss the CDCR tasks that each of the datasets creates for CDCR models, i.e., lexical disambiguation and lexical diversity challenges. Finally, we continue the discussion about the future of the CDCR evaluation started by \newcite{bugert2021generalizing} to evaluate the models on multiple event-CDCR datasets and propose a task-driven CDCR evaluation to aim at improved robustness of CDCR models \footnote{The datasets (CoNLL format for coreference resolution \cite{pradhan-etal-2012-conll}, mentions in JSON format \cite{barhom-etal-2019-revisiting,Eirew2021}) and the code for the statistics calculation is located in 
\url{https://github.com/anastasia-zhukova/Diverse_CDCR_datasets}.}.

\section{Related work}
\label{sec:rel}

Coreference resolution (CR) and cross-document coreference resolution (CDCR) are tasks that aim to resolve coreferential mentions in one or multiple documents, respectively \cite{Singh_2019}. (CD)CR approaches tend to depend on the annotation schemes of the CDCR datasets that specify a definition of mentions and coreferential relations \cite{bugert2021generalizing}. The established CDCR datasets are typically event-centric \cite{cybulska-vossen-2014-using,mitamura2015overview,ogorman-etal-2016-richer,hong-etal-2016-building,mitamura2017events,vossen-etal-2018-dont,BugertRBDG20,Eirew2021,bugert-gurevych-2021-event}, i.e., triggers annotating a mention at the presence of an event, or concept-centric \cite{weischedel2011ontonotes,recasens-etal-2010-typology,hasler2006nps,minard-etal-2016-meantime}, i.e., annotates mentions if an antecedent contains a minimum number of coreferential mentions of occurring entities or events in the documents.

Most (CD)CR datasets contain only strict identity relations, e.g., TAC KBP \cite{mitamura2017events,mitamura2015overview}, ACE \cite{linguistic2008ace,linguistic2005ace}, MEANTIME \cite{minard-etal-2016-meantime}, OntoNotes \cite{weischedel2011ontonotes}, ECB+ \cite{bejan2010unsupervised,cybulska-vossen-2014-using}, GVC \cite{vossen-etal-2018-dont}, FCC \cite{BugertRBDG20}, WEC \cite{Eirew2021}, and HyperCoref \cite{bugert-gurevych-2021-event}. Less commonly used (CD)CR datasets explore relations beyond strict identity. For example, NiDENT \cite{recasens-etal-2012-annotating} is a CDCR dataset of entities-only mentions that was created by reannotating NP4E. NiDENT explores coreferential mentions of more loose coreference relations coined near-identity that among all included metonymy, e.g., ``White House'' to refer to the US government, and meronymy, e.g., ``US president'' being a part of the US government and representing it. Reacher Event Description (RED), a dataset for CR, contains also more loose coreference relations among events \cite{ogorman-etal-2016-richer}. 

Mentions coreferential with more loose relations are harder to annotate and automatically resolve than mentions with identity relations \cite{recasens2010typology}. Bridging relations occur when a connection between mentions is implied but is not strict, e.g., a ``part-of'' relation. Bridging relations, unlike identity relations, form a link between nouns that do not match in grammatical constraints, i.e., gender and number agreement, and allow linking noun and verb mentions, thus, constructing abstract entities \cite{kobayashi-ng-2020-bridging}. The existing datasets for identification of bridging relations, e.g., ISNotes \cite{Hou2018a}, BASHI \cite{rosiger-2018-bashi}, ARRAU \cite{poesio-artstein-2008-anaphoric} annotate the relations only of noun phrases (NPs) on a single-document level and solve the problem as antecedent identification problem rather than identification of a set of coreferential anaphora \cite{Hou2018a}. GUM corpus \cite{Zeldes2016} annotates both coreference and bridging relations but as two separate tasks. Definition identification in the DEFT dataset \cite{spala-etal-2019-deft} focuses on annotating mentions that are linked with ``definition-like'' verb phrases (e.g., means, is, defines, etc.) but does not address linking the antecedents and definitions into the coreferential chains.

In the following sections, we discuss and contrast two perspectives on annotating coreferential relations in CDCR datasets: coreferential relations of identity strength in event-centric chains of mentions and more loose coreference relations, i.e., a combination of identity and bridging anaphoric relations such as synonymy, metonymy, meronymy \cite{Poesio1998,rosiger-2018-bashi}, in concept-centric chains. We aim at exploring if loose coreference relations are a challenge to the CDCR task and if it is possible to solve with feature-based approaches.

\section{Coreference anaphora in CDCR}
\label{sec:dataset}
We compare ECB+ \cite{cybulska-vossen-2014-using}, a state-of-the-art CDCR dataset, and NewsWCL50 \cite{Hamborg2019a}, a dataset that contains annotations of frequently reported concepts that tend to contain biased phrases by word choice and labeling in news articles. The two datasets annotate both event and entity mentions but significantly differ in how each defines coreference chains. To our knowledge, despite not being primarily designed as a CDCR dataset, NewsWCL50 is the only CDCR-related dataset that annotated coreferential chains with identity and looser relation together in NPs and verb phrases (VPs).



\subsection{Qualitative comparison}  
We qualitatively compare the datasets regarding three factors that represent the main characteristics in CDCR. First, properties of the topic composition of the CDCR datasets. Second, how each dataset defines which phrases should become coreference mentions, i.e., candidates for a single coreference chain, and how these phrases should be annotated. Third, how coreferential chains are constructed, i.e., which relations link the coreferential anaphora. To conclude the qualitative analysis, we perform a small reannotation experiment and take a few exemplary entities and events from ECB+ and annotate mentions and coreference chains for them following a coding book of NewsWCL50.

\subsubsection{Dataset structure}
CDCR is performed on a collection of narrowly related articles based on topic or event similarity. 

NewsWCL50 contains one level of topically related news articles, i.e., articles reporting on the same event. Although reporting on the same event, each news article discusses at least partially different aspects of the event. For example, in one of NewsWCL50’s topics (topic \#5), while generally reporting on Trump visiting the UK, some news articles focused on Trump's plans during this visit, some compared his visit to the other visits of the US presidents, and others on the demonstrators opposing this visit.

ECB+ contains news articles related on two levels: first, on a general topic level of one event, e.g., an earthquake, and second, on a subtopic level of an event described by an action, actor(s), location, and time. Therefore, CDCR on ECB+ is possible on a topic or sub-topic levels \cite{cybulska-vossen-2014-using,upadhyay-etal-2016-revisiting,Cattan2021a}. The authors later validated the dataset and provided a list of these validated documents and sentences \cite{cybulska2015bag} that we now reuse to report about ECB+.

For ECB+, \newcite{cybulska2015bag} proposed a split into train, validation, and test subsets. On the contrary, \newcite{Hamborg2019a} used the entire corpus of NewsWCL50 as a test set for their unsupervised approach due to the small number of annotated topics (see \Cref{tab:comp}).


\begin{table*}
\caption{Comparison of exemplary mentions annotated chains in ECB+ and when re-annotated as concepts following the NewsWCL50 coding book.}
\small
\centering
\begin{tabular}{p{0.06\textwidth}|p{0.06\textwidth}|p{0.8\textwidth}}
\hline
\textbf{Chain's name} & \textbf{Coding book} & \makecell[c]{\textbf{Annotated mentions}}\\
\hline

    \makecell[c]{\multirow{2}{*}{\rotatebox[origin=r]{90}{\textbf{36: Warren Jeffs} (entity)}}} & \makecell[c]{\multirow{1}{*}{\rotatebox[origin=r]{90}{\textbf{ECB+}}}   }               & \textit{t36\_warren\_jeffs}: attorney; FLDS leader's; he; head; him; his; Jeffs; leader; leader Warren Jeffs; pedophile; Polygamist; polygamist leader Warren Jeffs; Polygamist prophet Warren Jeffs; polygamist sect leader Warren Jeffs; Polygamist Warren Jeffs; Warren Jeffs; Warren Jeffs, Polygamist Leader; who \\ 
                        \cline{2-3} 
                       & \makecell[c]{\multirow{1}{*}{\rotatebox[origin=r]{90}{\textbf{NewsWCL50}}}  }              &  a handful from day one; a problem; a victim of religious persecution; an accomplice for his role; an accomplice to rape by performing a marriage involving an underage girl; an accomplice to sexual conduct with a minor; an accomplice to sexual misconduct with minors; an accomplice to the rape of a 14-year-old girl; FLDS prophet Warren Jeffs; God’s spokesman on earth; her father; his client; Jeffs;  Jeffs, 54; Jeffs, who acted as his own attorney; Jeffs, who was indicted more than two years ago; Mr. Jeffs; one individual, Warren Steed Jeffs; one of the most wicked men on the face of the earth since the days of Father Adam; penitent; Polygamist prophet Warren Jeffs; polygamist sect leader Warren Jeffs; polygamist Warren Jeffs; president; prophet of the Fundamentalist Church of the Jesus Christ of the Latter Day Saints; prophet Warren Jeffs; stone-faced; The 54-year-old Jeffs; the defendant; the ecclesiastical head of the Fundamentalist Church of Jesus Christ of Latter Day Saints; the father of a 15-year-old FLSD member's child; the highest-profile defendant; the prophet; the self-styled prophet; their client; their spiritual leader; this individual; Warren Jeffs; Warren Jeffs, leader of the Fundamentalist Church of Jesus Christ of Latter Day Saints; Warren Jeffs, polygamist leader \\ 
\hline

 \makecell[c]{\multirow{4}{*}{\rotatebox[origin=r]{90}{\textbf{39: Become new Dr.Who}    (event)}}} & \makecell[c]{\multirow{3}{*}{\rotatebox[origin=c]{90}{\textbf{ECB+}}}} & 
                        \textit{t39\_capaldi\_play\_doc}: play, take on, take up \\
                         \cline{3-3} 
                       &  & \textit{t39\_replace\_smith}: replace, replacing, stepped into, take over, takes over \\ 
                       \cline{3-3} 
                       & & \textit{t39\_play\_doc}: one, play, role \\

                       \cline{2-3} 
                       &  \makecell[c]{\multirow{1}{*}{\rotatebox[origin=r]{90}{\textbf{NewsWCL50}}}}               & about to play the best part on television; an incredible incarnation of number 12; become one of the all-time classic Doctors; become the next Doctor Who; becoming the 12th Doctor; becoming the next Doctor; Being asked to play The Doctor; being the doctor; had been chosen; get started; getting it; has been announced as the new star of BBC sci-fi series Doctor Who; has been named as the 12th actor to play the Doctor; his unique take on the Doctor; is about to play the best part on television; is to be the new star of Doctor Who; might be the right person to take on this iconic part; play it; play the Doctor; revealed as 12th Doctor; stepped into Matt Smith's soon to be vacant Doctor Who shoes; take over from Matt Smith as Doctor Who; take the role for days; takes over Doctor Who Tardis; the Doctor's appointment; replace Matt Smith on Doctor Who;  will be the 12th actor to play the Doctor; will play the 12th Doctor; will replace Matt Smith; will take over from Smith \\ 
\hline

\end{tabular}
\label{tab:annot_example}
\end{table*}

\subsubsection{Mentions}
In the ECB+ dataset, the annotation process is centered around events as main concept-holders. Four components describe each event, i.e., action, time, location, and participant \cite{cybulska-vossen-2014-using}. Due to its focus on events, NP and VP phrases are annotated as mentions only if defining an event, resulting later in (1) many in-text mentions of entities not being annotated and, thus, not being part of the dataset’s annotated coreference chains or new non-event-centric coreference chains, (2) many action or entity singleton-mentions that occur only once. For human and non-human participants ECB+ allows annotating pronouns. 

In contrast, in NewsWCL50, all mentions are annotated if at least five candidate phrases occur in at least one document in NewsWCL50 \cite{Hamborg2019a}. The goal is to identify the most frequently reported concepts within the related news articles. NewsWCL50 annotates mentions of NPs and VPs. The annotation scheme distinguishes the mention chains referring to actors, actions, objects, events, geo-political entities (GPE), and abstract entities \cite{poesio-artstein-2008-anaphoric}\footnote{Event mentions are identified in ECB+ as those that belong to the types ``action'' and ``negative action.'' In NewsWCL50, the event mention types belong to ``action,'' ``event,'' and ``misc.''}. NewsWCL50 excluded annotating any pronouns. Although, the authors stated that they annotated a concept if a number of mentions reached a certain amount, we identified a few inconsistencies across the dataset. For example, some GPEs we annotated with fewer mentions or multiple frequently occurring concepts with 5-10 mentions were not annotated. Moreover, concepts that refer to date-time and locations (if not referring to GPEs) are excluded from annotations. 

In both datasets, the mentions length defines which tokens of a phrase are annotated as a mention. Depending on a mention’s type in ECB+, a mention is annotated with a ``maximum’’ or ``minimum span’’ style. Mentions of location and time are annotated with a ``maximum span’’ style and contain all words that belong to a mention phrase. On the contrary, for the mentions of human, non-human participants ECB+ demands a ``minimum span’’ style, i.e., phrases contain the smallest number of words that preserve theier core meaning. Often this annotation style leads to mentions that consist only the heads of phrases. The NewsWCL50 annotation scheme follows a ``maximum span’’ style for all mentions and yields longer phrases. ``A maximum span’’ style allows capturing modifiers of the heads of phrases and the overall meaning that can change with the modifies, e.g., compare ``kids'' to ``undocumented \textit{kids} who did nothing wrong'' and ``American \textit{kids} of undocumented immigrant parents.''  ECB+ focuses on identifying the most reliable spans either if they are to be annotated or automatically extracted, whereas NewsWCL50 focuses on annotating the longest coreferential phrase as the determining words my carry the bias by word choice and labeling.


\subsubsection{Relations}
The annotation schemes differ in the strength of the coreferential relations that link the mentions, i.e., identity or more loose relations that include bridging and near-identity. 


ECB+ links NP-mentions within and across the documents if these mentions refer to the same participant, location, or time, e.g., ``the U.S.’’ -- ``America’’ refer to the same entity across multiple documents. ECB+ links the action mentions into one chain if (1) the mentions are related with strict identity, (2) VP-mentions belong to the same event, i.e., have similar attributes of participant(s), location, and time (could be verbal or nominal mention) \cite{cybulska2014guidelines}. Therefore, if an article that reports on immigration to the US covers either two points in time within a couple of weeks difference or if the immigrants are located on multiple spots on the immigration way, annotators will need to create various separate entities for the entity ``immigrants.'' 

NewsWCL50 links the mentions if they are coreferential with a mix of identity and bridging relations \cite{Poesio1998}: (1) mentions refer to the same entity/event including subject-predicate coreference (e.g., ``Donald Trump'' -- ``the newly-minted president''), (2) synonym relations (e.g., ``a caravan of immigrants'' -- ``a few hundred asylum seekers''), (3) metonymy (e.g., ``the Russian government'' -- ``the Kremlin''), (4) metonymy/holonymy (e.g., ``the U.S.-Mexico border'' -- ``the San Ysidro port of entry''), (5) mentions are linked with copular verbs, e.g., ``be’’ or ``seem’’ or other phrases that establish association, such as ``call’’ in ``Trump called Kim Jong Um a Little Rocket Man’’ (e.g., ``Kim Jong Un'' -- ``Little Rocket Man'' or ``crossing into the U.S.'' -- ``a crisis at the border''), (6) mentions meet GPE definitions as of ACE annotation \cite{linguistic2008ace} (e.g., ``the U.S.'' -- ``American people'), (7) mentions are elements of one set (e.g., ``guaranteed to disrupt trade'' -- ``drove down steel prices'' -- ``increased frictions'' as members of a set ``Consequences of tariff imposition''). The annotation scheme requires annotating the generic use of mentions with ``part-of'' relations referring to GPEs. For example, annotators must always annotate mentions of the police as referring to the U.S. because the biased language used to report about the police can affect the perception of the U.S. in general \cite{hamborg2019codebook}.

\subsubsection{Reannotation experiment}
\label{sec:reann}
To provide a qualitative example of the differences of both annotations schemes, we reannotated an entity and events of the state-of-the-art ECB+ by following a more experimental coding book of NewsWCL50 \cite{hamborg2019codebook}. The goal is to illustrate how coreference chains may differ if mentions not only with strict identity relations were linked.

\Cref{tab:annot_example} shows an example of how coreference chains of the two annotation schemes differ. The table shows how the exemplary mentions are annotated in ECB+ and NewsWCL50. The NewsWCL50 coding book yields coreference chains that (1) annotate any occurred coreferential mentions in the text, not only related to an event, (2) contain a mix of strict identity and loose bridging coreferential anaphora. 

The reannotated ECB+ entities contain more mentions compared to the original ECB+ annotation. ECB+ annotated entities in an event-centric way, whereas NewsWCL50 annotated any occurrence of the mentions referring to the same entity/event. Moreover, NewsWCL50's coding book annotates coreferential mentions that were used to describe an entity via association with this entity, e.g., ``Warren Jeffs'' -- ``God’s spokesman on earth'' or ``become one of the all-time classic Doctors'' -- ``an incredible incarnation of number 12.'' 

\subsubsection{Summary} ECB+ annotations cover narrowly defined entities that describe an event as precisely as possible regarding the event's participants, action, time, and location. In contrast, the annotation scheme employed for the creation of NewsWCL50 aims at determining broader coreference chains that include mentions of entities that would have not been annotated or would be split into multiple chains if the ECB+ annotation scheme was used.

\subsection{Quantitative comparison}
\label{sec:ld}
Besides the previously discussed qualitative differences, we also quantitatively compare the annotation schemes. We compare the lexical diversity of coreference chains, their sizes, numerical parameters of the corresponding datasets, and annotation complexity measured by inter-coder reliability.

\begin{table}[h]
\caption{Quantitative comparison of ECB+ and NewsWCL50. Both datasets were parsed from scratch from the original annotation files.}
\label{tab:datasets}
\centering
\small
\begin{tabular}{l|c  c }
\hline
\textbf{Criteria} & \textbf{ECB+} & \textbf{NewsWCL50} \\
\hline
\# topics & 43 & 10 \\ 
\# subtopics & 86 & - \\
\# articles & 962 & 50  \\ 
\# tokens & 377 367  & 49 591  \\ 
\# mentions & 12 004 & 5600  \\ 
\# event mentions & 4 013  & 1225  \\
\# entity mentions & 7 991  & 4375  \\
\# chains & 4 759 & 170  \\
\# singletons & 3 445 & 10 \\
inter-coder reliability (ICR) & 0.76 & 0.65 \\
\hline
\textbf{with singletons} & \\
average chain size & 2.5 & 32.9   \\
F1$_{CoNLL}$ & 72.9 & 46.8 \\
average \# unique lemmas & 1.4 & 9.9 \\
phrasing diversity (PD) & 1.3  & 9.7  \\
\hline
\textbf{without singletons} & \\
average chain size & 6.4 & 35.0   \\
F1$_{CoNLL}$ & 64.4 & 46.5 \\
average \# unique lemmas & 2.4 & 10.5 \\
phrasing diversity (PD) & 1.4  & 9.7  \\
\hline
\end{tabular}
\label{tab:comp}
\end{table}

\subsubsection{Lexical diversity}
Lexical diversity indicates how the wording differs within annotated concepts and how obvious or hard is the task of resolving such mentions. We evaluate lexical diversity with three metrics: (1) CoNLL F1 score of a primitive same-head-lemma baseline \cite{cybulska-vossen-2014-using,pradhan-etal-2012-conll} and (2) the number of different head-lemmas in coreference chains \cite{Eirew2021}, (3) a new metric for lexical diversity that accounts for more within-chain phrase variations than a number of unique lemmas. \newcite{Cattan2021a} revealed that the performance of CDCR models using the original version of ECB+ with singletons facilitated higher CoNLL F1 score, thus, making CDCR task look easier than it is. To adhere to this strategy of fair results, we calculate all metrics for lexical diversity on the annotated chains that exclude singletons. \Cref{tab:comp} summarizes all metrics and presents general statistics of the datasets, e.g., a number of annotated mentions and coreference chains. 

\subsubsection*{Accuracy of a lemma CDCR baseline} A CDCR baseline method is to resolve mentions with same-lemmas within related sets of documents. The accuracy of a lemma-baseline shows that the better this method performs, the lower is the lexical variation \cite{cybulska-vossen-2014-using}. \Cref{tab:comp} reports that same-lemmas is more efficient on the chains in the ECB+ dataset and denotes smaller lexical diversity of the mentions in ECB+ compared to NewsWCL50. Coreferential anaphora with more loose relations increases the possible number combinations of the predicted chains, thus, increases the complexity of the CDCR task \cite{kobayashi-ng-2020-bridging}. 

\subsubsection*{Number of unique lemmas of phrases' heads} \newcite{Eirew2021} proposed to measure lexical diversity with an average number of unique lemmas of phrases' heads in coreference chains. Similar to $F1_{CoNLL}$ on the lemma-baseline, the number of unique lemmas shows that  NewsWCL50 contains almost four times more diverse coreference chains than ECB+ (see \Cref{tab:comp}). 

\begin{figure}[h]
\centering
\includegraphics[width=0.49\textwidth]{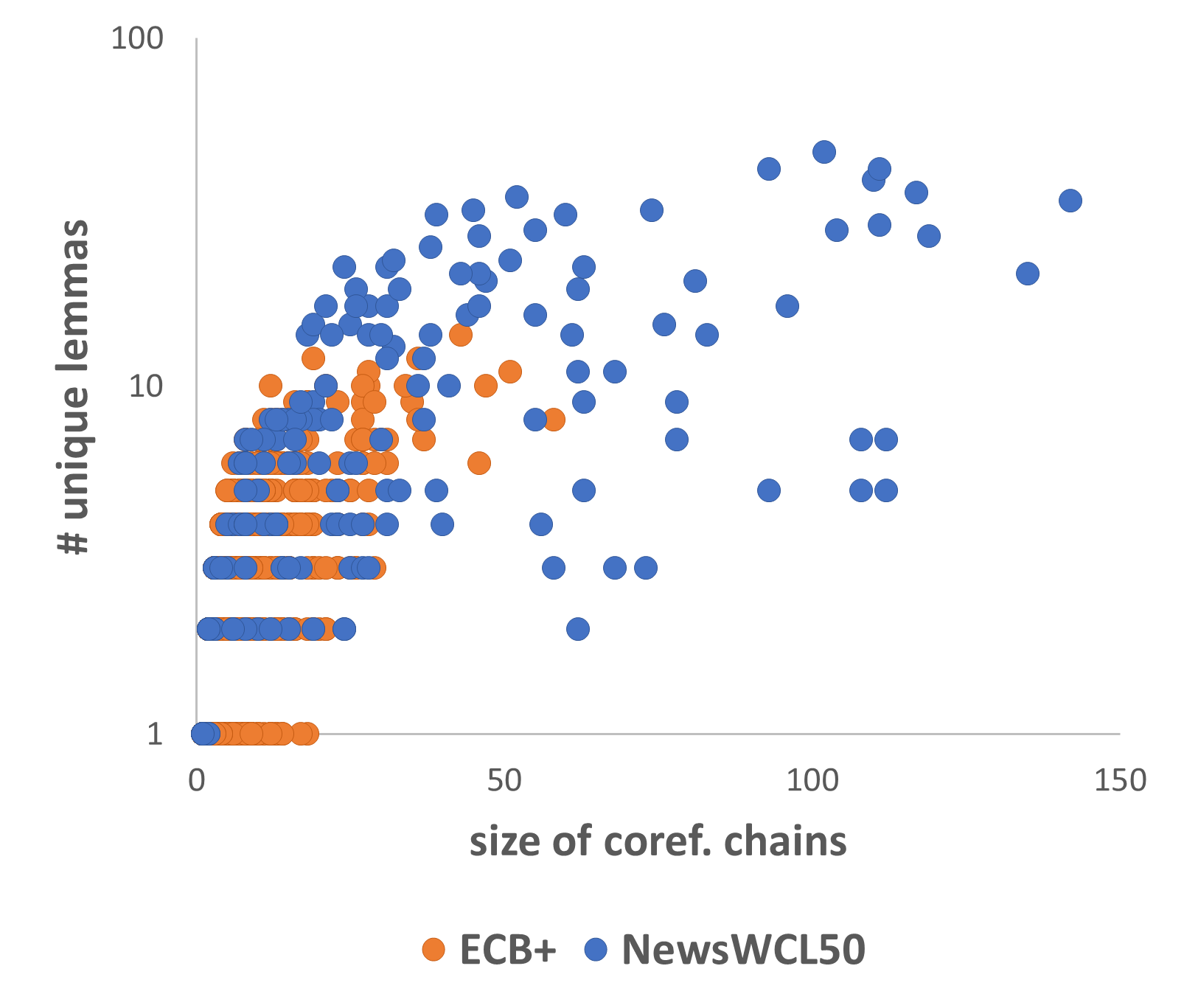}
\caption{Very subtle difference in lexical diversity between ECB+ and NewsWCL50 datasets when the diversity is measured with unique lemmas per coreference chain as proposed by Eirew et al. (2021). The Y-axis in logarithmic scale.}
\label{img:un_lemmas}
\end{figure}

Unique lemmas as a metric for lexical diversity used in \newcite{Eirew2021} has the most advantage in a ``minimum span'' mention annotation style, i.e., annotate mainly heads of phrases. Unfortunately, this metric does not account for larger phrasing diversity in a ``maximum span'' style, i.e., the longest spans' annotation that includes all head modifiers instead of only minimum descriptive ones. \Cref{img:un_lemmas} depicts the trends in the two annotation styles where numbers of the unique lemmas in the coreference chains (logarithmic scale) are plotted against the sizes of the chains. Although the average value of NewsWCL50 is higher than ECB+, the trends suggest that larger subtopics of many documents annotated with ECB+ coding book would yield similarly diverse coreference chains as NewsWCL50.

\begin{figure*}[h]
\centering
\includegraphics[width=1\textwidth]{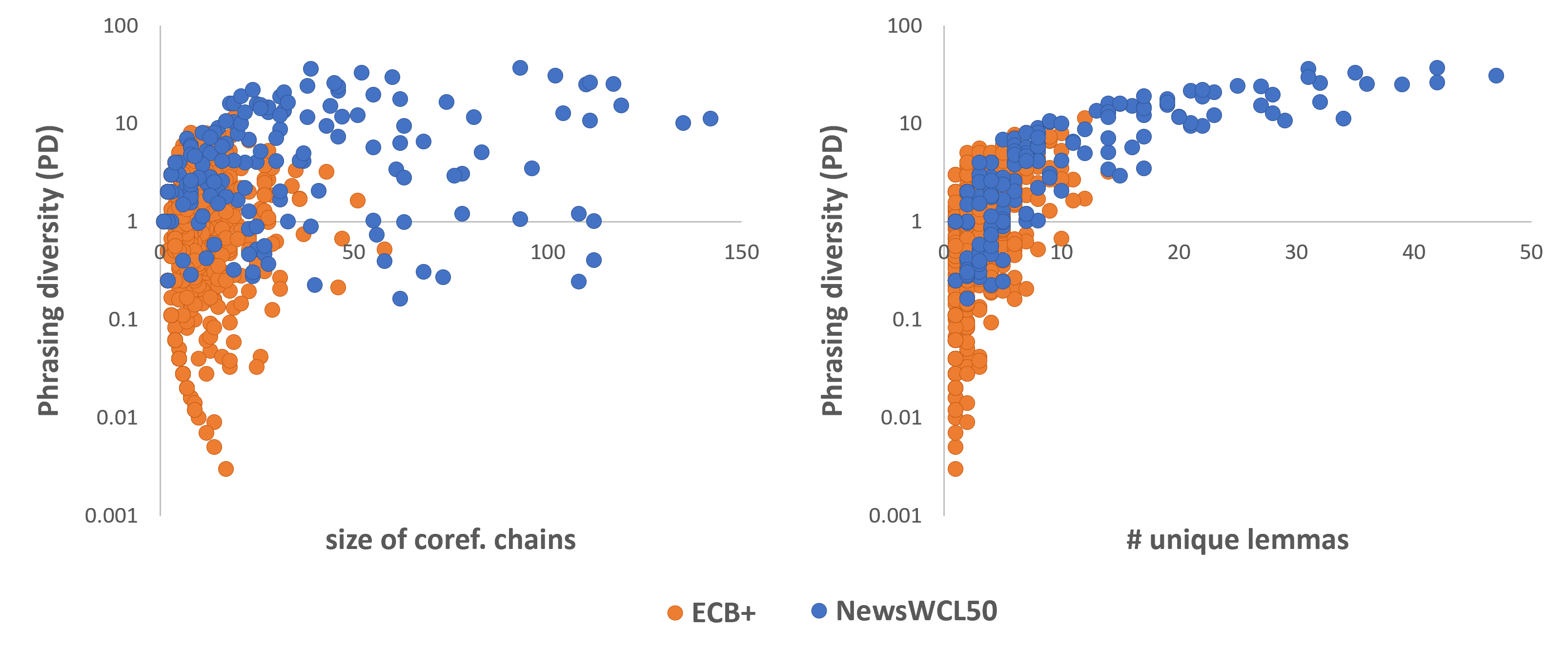}
\caption{Comparison of coreference chains from ECB+ and NewsWCL50 with a phrasing diversity metric (PD): the difference between the coreference chains is more clear between the chains of the same size and with the same number of unique lemmas. The lexical diversity of NewsWCL50 more clearly visible to be higher than ECB+. Y-axes of PD are in logarithmic scale.}
\label{img:comp}
\end{figure*}

\subsubsection*{Phrasing diversity metric} To encounter for the lexical diversity beyond unique lemmas, we introduce a new \textit{phrasing diversity metric} (PD) that represents the variation of complete phrases used to refer to the same entity or event. PD measures lexical variation of coreference chains using the mention frequency and phrasing diversity: 
\begin{align}
PD_c = \begin{cases}
\frac{\sum_{\forall h \in H_c}\frac{|U_h|}{|P_h|} \cdot \sum_{\forall h \in H_c} |U_h|}{|M_c|}, &  \text{ if }  |M_c| > 1 \\
1, & \text{else}
\end{cases}
\end{align}
where $H_c$ represents all unique \textit{heads} of phrases of an annotated chain $c$, $|U_h|$ is the number of \textit{unique phrases} with head $h$, $|P_h|$ is the number of \textit{all phrases} with head $h$, and $|M_c|$ is a number of \textit{all mentions} of an annotated chain $c$. Lastly, to aggregate PD per dataset, we compute weighted average over all corresponding chains: 
\[PD = \frac{\sum_{\forall c \in C} |M_c| \cdot PD_c}{\sum_{\forall c \in C} |M_c|}\]



Similar to \Cref{img:un_lemmas}, \Cref{img:comp} plots PD against the sizes of coreference chains for each dataset and shows the higher lexical diversity of NewsWCL50 compared to ECB+ (Y-axis is in logarithmic scale). The trends suggest that coreference chains of ECB+ are narrowly defined: the large coreference chains have lower values of PD than NewsWCL50, i.e., the chains contain a lot of identical mentions. The figure shows that a large majority of the ECB+ concepts are located around value of 1, whereas NewsWCL50's concepts are distributed mostly above the value of 1. Unlike ECB+, coreference chains of NewsWCL50 are more broadly defined, i.e., have large values of both PD and an average size of coreference chains (see \Cref{tab:comp}). This finding supports the qualitative example in \Cref{sec:reann} where reannotating leads to the increased size of original ECB+ chains due to (1) annotated mentions with more loose coreference relations, (2) merging previously annotated chains of ECB+ leading to more broadly-defined chains, (3) the annotating maximum spans of mentions that capture the lexical complexity of mentions. In ECB+, some large coreference chains get lower PD value due to largely repetitive heads of mentions in these chains.  

\begin{table*}[]
\centering
\caption{A comparison of two metrics of lexical diversity: unique lemmas and phrasing diversity (PD). While the numbers of unique lemmas are identical for the two chains in the example, PD accounts more for the variation in the phrases.
}
\small
\begin{tabular}{c|l|c|c|c}
& Mentions & \# repetitions   & \makecell[c]{\# unique  lemmas}     & \makecell[c]{Phrasing diversity (PD)} \\
\hline
  \multirow{4}{*}{\rotatebox[origin=c]{90}{chain 1}} 
   & Donald \textit{Trump} &  3 & \multirow{5}{*}{2}& \multirow{4}{*}{1.1}\\
    \cline{2-3}
   & the \textit{president}   & 1 &  & \\
   \cline{2-3}
   & \makecell[l]{President Donald  \textit{Trump} }  & 1 & & \\
   \cline{2-3}
   & Mr. \textit{Trump}   & 1 & & \\
    \hline
    \multirow{6}{*}{\rotatebox[origin=c]{90}{chain 2}} 
    &  \makecell[l]{undocumented  \textit{immigrants}} & 1 &  \multirow{6}{*}{2} &   \multirow{6}{*}{2.0}\\
    \cline{2-3}
    & \makecell[l]{\textit{immigrants}  seeking hope} & 1 & & \\
    \cline{2-3}
    & \makecell[l]{unauthorized  \textit{immigrants} }& 1 &  & \\
    \cline{2-3}
    & migrant \textit{caravan} & 1 & & \\
    \cline{2-3}
    & \makecell[l]{a \textit{caravan}  of Central American  migrants} & 1 & & \\
    \cline{2-3}
    & \makecell[l]{a \textit{caravan}  of hundreds of migrants} & 1 & & \\

\hline
\end{tabular}
\label{tab:ld_example}
\end{table*}

\Cref{img:comp}b shows the difference between unique lemmas and the proposed PD metric, i.e., the same unique lemmas have higher or lower PD depending on the fraction of repetitive phrases to be resolved. Consider the example in \Cref{tab:ld_example}: 

\[PD_{chain 1} = \frac{(3/5 + 1/1)\cdot(3 + 1)}{6} = 1.1\]
\[PD_{chain 2} = \frac{(3/3 + 3/3)\cdot(3 + 3)}{6} = 2.0\]

The example shows that for the same number of unique lemmas, a PD metric may differ and show higher or lower variation depending on the number of unique head lemmas and phrases with these heads. \Cref{img:comp}b depicts that NewsWCL50 annotates chains with larger PD for the same number of unique lemmas compared to ECB+. Therefore, PD is capable of indicating larger lexical variation than unique lemmas.

\subsubsection{Inter-coder reliability}
We explore inter-coder reliability (ICR). For the strictly formalized ECB+, \newcite{cybulska-vossen-2014-using} reported Cohen's Kappa for the annotation of the event/entity mentions $\kappa_C=0.74$ and for connecting these mentions into coreferential chains $\kappa_C=0.76$. For the more flexible NewsWCL50 scheme,  \newcite{Hamborg2019a} reported an average observed agreement $AOA=0.65$ \cite{BYRT1993423}. While these measures cannot be compared directly, they indicate that schemes including mentions with more loosely related anaphora may result in lower ICR since AOA is considered less restrictive than Kappa \cite{recasens-etal-2012-annotating,BYRT1993423}. Moreover, a lower ICR indicates that the annotation of NewsWCL50 tends to be more subjective and requires annotators to undergo specific training and be familiar with a domain of political news, e.g., have a background in political science or journalism.

\subsubsection{Summary} 
\label{sec:sum_quant}
\Cref{tab:comp} reports numeric properties, i.e., the number of annotated documents and mentions, and quantitative description parameters of the datasets, i.e., lexical diversity. On the one hand, ECB+ includes more topics and articles than NewsWCL50 (43/962 against 10/50) and annotates two times more mentions (12004 and 5600). On the other hand, the density of annotations per article is higher in NewsWCL50, i.e., 112 annotations per article in NewsWCL50 against 13 in ECB+, which shows longer documents in NewsWCL50 and more thoroughly covered annotation of mentions. Moreover, annotation rules of NewsWCL50 avoid creating singletons (5.9\% of the chains are singletons) whereas 72.3\% of the chains from ECB+ are singletons. \newcite{Cattan2021a} showed that evaluation metrics on the datasets without singleton exclusion artificially inflates performance of the CDCR models.

\begin{figure}[h]
\centering
\includegraphics[width=0.49\textwidth]{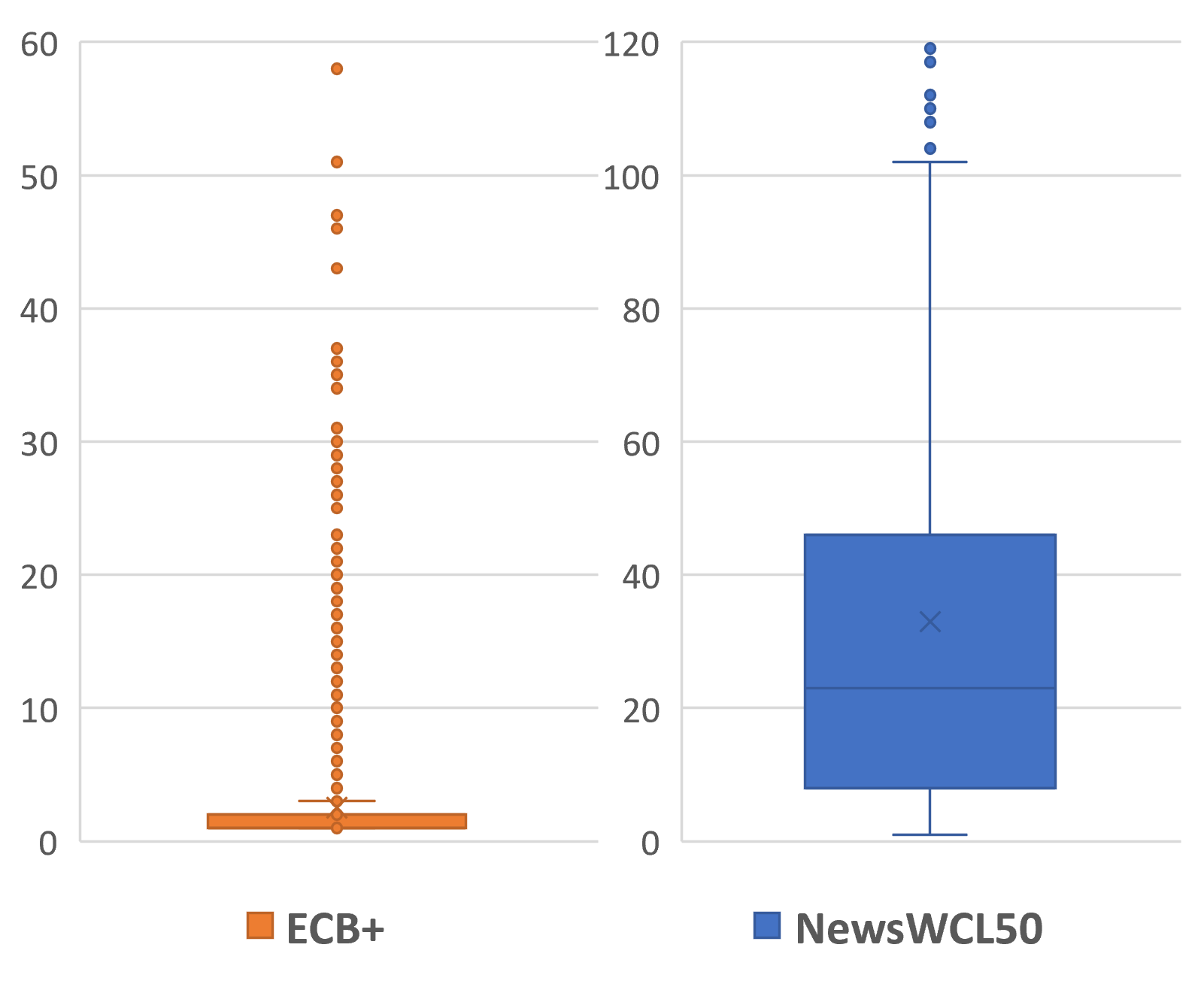}
\caption{Difference in the size of coreference chains between ECB+ and NewsWCL50: ECB+ focuses on annotating some coreferential cases of strict identity relations whereas NewsWCL50 aims at covering all coreferential mentions with diverse identity relations.}
\label{img:mentions}
\end{figure}

We evaluated the complexity of the coreference chains with the average number of contained mentions, i.e., the size of the coreference chains, and multiple metrics of lexical diversity. To ensure a fair comparison, we calculated datasets' parameters on both present and removed singletons chains. Without singletons, the average number of mentions per coreference chain is more than five times larger in NewsWCL50 than in ECB+. \Cref{tab:comp} shows that lexical diversity of ECB+ measured by both unique lemmas and PD is almost 4-5 times lower than those of NewsWCL50. Moreover, NewsWCL50 has a higher value of $F1_{CoNLL}$ on the head-lemma baseline on a subtopic level, thus, indicating a more complex task for a CDCR model. The comparison of the inter-annotator reliability shows that annotation of a dataset following a NewsWCL50 coding book can lead to more subjective results than annotating following ECB+. \Cref{img:mentions} depicts the resulting difference between the sizes of coreference chains in ECB+ and NewsWCL50.

\section{Discussion}

In our quantitative and qualitative analysis, ECB+ showed the smaller average size of coreference chains and the lower lexical diversity measured by all three metrics, e.g., $F1_{CoNLL}$ on a primitive CDCR method, number of unique lemmas of phrases' heads, and phrasing diversity metric (PD). NewsWCL50 shows that annotating both identity, synonym, metonymy/meronymy, bridging, and subject-predicate coreference relations yields an increase of the lexical diversity in the annotated chains (see \Cref{tab:comp}). Consequently, the increased lexical diversity results in a higher level of semantic complexity and abstractness in the annotated coreferential chains, i.e., the resolution of some mentions requires an understanding of the context in which they appear. Annotating NewsWCL50 leads to lower inter-coder reliability compared to ECB+ and indicates of a  more challenging task for humans to capture looser coreference relations. The increase in the lexical diversity and the abstractness of coreference chains thus poses a new challenge to the established CDCR models. 

Although ECB+ annotates narrowly-defined coreference chains with smaller lexical diversity on a subtopic level, ECB+ creates a lexical ambiguity challenge, i.e., CDCR models need to resolve mentions on the mixed documents of two subtopics, which contain verbs that refer to different events \cite{cybulska-vossen-2014-using,upadhyay-etal-2016-revisiting,Cattan2021a}. Therefore, ECB+ and NewsWCL50 establish two diverse CDCR tasks: lexical ambiguity and lexical diversity.

The diversity of the annotation schemes supports the conclusion by \newcite{bugert2021generalizing} that CDCR models need to be evaluated on multiple CDCR datasets. \newcite{bugert2021generalizing} evaluated the state-of-the-art CDCR models on multiple event-centric datasets and reported their unstable performance across these datasets. The authors suggested evaluating every CDCR model on four event-centric datasets and report the performance on a single-document, within-subtopic, and within-topic levels. 


For further CDCR evaluation, we see a need to evaluate CDCR models on the two tasks: lexical ambiguity and lexical diversity. Evaluation should consist of multiple event-centric datasets, which focus on disambiguating events and subtopic, and include concept-centric datasets, which focus on resolving mentions with a mix of loose context-specific anaphoric and strict identity coreference relations. Such a setup with diverse CDCR datasets will facilitate the fair and comparable evaluation of coreferential chains of various natures and exposes models to the challenges of different nature.

\section{Conclusion}
CDCR research has a long history of focusing on the resolution of mentions annotated in an event-centric way, i.e., the occurrence of events triggers the annotation of mentions of these events and entities as their attributes. We compared a state-of-the-art event-centric CDCR dataset, i.e., ECB+, to the concept-centric NewsWCL50 dataset that explored the identification of high variance in news articles as a CDCR task. The reviewed CDCR datasets reveal a large scope of relations that define coreferential anaphora, i.e., from strict identity in ECB+ to mixed identity and bridging of coreference relations in NewsWCL50. 

To ensure CDCR can robustly handle data variations on the lexical disambiguation challenge, \newcite{bugert2021generalizing} proposed evaluating CDCR models on three event-centric datasets on multiple levels of coreference resolution, i.e., a document, subtopic, and topic levels. We proposed a phrasing diversity metric (PD) that enabled a more fine-grained comparison of lexical diversity in coreference chains than the average number of unique head lemmas \cite{Eirew2021}. We show that CDCR datasets can contain coreference chains with higher lexical diversity than more common event-CDCR dataset. High phrasing diversity metric together with low inter-coder reliability indicate that difficulty of annotating of loose coreference relations also poses a new challenge in the CDCR research. We propose to create an additional CDCR challenge, i.e., lexical diversity challenge, and include CDCR datasets that annotate coreference chains with high lexical variance chains, e.g., NewsWCL50. 

\section{Bibliographical References}\label{reference}

\bibliographystyle{lrec2022-bib}
\bibliography{lrec2022-example}

\bibliographylanguageresource{languageresource}

\end{document}